\documentclass{article}

\usepackage[final]{nips_2016}
\setcitestyle{numbers,square,citesep={,},aysep={,},yysep={,}}

\usepackage[utf8]{inputenc} 
\usepackage[T1]{fontenc}    
\usepackage{hyperref}       
\usepackage{url}            
\usepackage{booktabs}       
\usepackage{amsfonts}       
\usepackage{nicefrac}       
\usepackage{microtype}      

\usepackage{times}
\usepackage{graphicx} 
\usepackage{caption}
\usepackage{subcaption}
\usepackage{wrapfig}

\usepackage{algorithm}
\usepackage{algorithmic}

\usepackage{hyperref}

\usepackage{amsmath}
\usepackage{amssymb}

\usepackage[font=footnotesize]{caption,subcaption}
\usepackage{sidecap}

\usepackage{xspace}

\usepackage{multirow}

\usepackage{color}

\newtheorem{lemma}{Lemma}[section]

\newcommand{\pushright}[1]{\ifmeasuring@#1\else\omit\hfill$\displaystyle#1$\fi\ignorespaces}

\long\def\ignorethis#1{}

\newcommand{\tr}{^\mathrm{T}}
\newcommand{\inv}{^{-1}}

\newcommand{\gauss}{\mathcal{N}}

\newcommand{\trace}{\text{tr}}

\newcommand{\trajdist}{p}
\newcommand{\policy}{\pi}

\newcommand{\params}{\theta}

\newcommand{\cost}{\ell}
\newcommand{\state}{\mathbf{x}}
\newcommand{\action}{\mathbf{u}}

\newcommand{\traj}{\tau}

\newcommand{\trajmu}{\mu^\trajdist}

\newcommand{\polmu}{\mu^\policy}
\newcommand{\polsig}{\Sigma^\policy}

\newcommand{\ucovar}{\mathbf{C}}

\newcommand{\ucovart}{\mathbf{C}_t}

\newcommand{\obj}{\mathcal{L}}

\newcommand{\kl}{D_\text{KL}}
\newcommand{\tv}{D_\text{TV}}
\newcommand{\ent}{\mathcal{H}}

\newcommand{\fct}{f_{c t}}
\newcommand{\fxt}{f_{\state t}}
\newcommand{\fut}{f_{\action t}}

\newcommand{\Qxt}{Q_{\state t}}
\newcommand{\Qut}{Q_{\action t}}

\newcommand{\Qxxt}{Q_{\state,\state t}}
\newcommand{\Quut}{Q_{\action,\action t}}
\newcommand{\Quxt}{Q_{\action,\state t}}

\newcommand{\kpol}{\mathbf{k}}
\newcommand{\Kpol}{\mathbf{K}}

\newcommand{\ft}{f_t}
\newcommand{\noise}{\mathbf{F}}

\newcommand{\st}{\state_t}
\newcommand{\at}{\action_t}

\title{Guided Policy Search as Approximate Mirror Descent}

\author{
  William Montgomery \\
  Dept. of Computer Science and Engineering\\
  University of Washington \\
  \texttt{wmonty@cs.washington.edu} \\
  \And
  Sergey Levine \\
  Dept. of Computer Science and Engineering\\
  University of Washington \\
  \texttt{svlevine@cs.washington.edu} \\
}

\begin{document}

\maketitle


\begin{abstract} 

Guided policy search algorithms can be used to optimize complex nonlinear policies, such as deep neural networks, without directly computing policy gradients in the high-dimensional parameter space. Instead, these methods use supervised learning to train the policy to mimic a ``teacher'' algorithm, such as a trajectory optimizer or a trajectory-centric reinforcement learning method. Guided policy search methods provide asymptotic local convergence guarantees by construction, but it is not clear how much the policy improves within a small, finite number of iterations. We show that guided policy search algorithms can be interpreted as an approximate variant of mirror descent, where the projection onto the constraint manifold is not exact. We derive a new guided policy search algorithm that is simpler and provides appealing improvement and convergence guarantees in simplified convex and linear settings, and show that in the more general nonlinear setting, the error in the projection step can be bounded. We provide empirical results on several simulated robotic navigation and manipulation tasks that show that our method is stable and achieves similar or better performance when compared to prior guided policy search methods, with a simpler formulation and fewer hyperparameters.

\end{abstract} 


\section{Introduction}
\label{sec:intro}

Policy search algorithms based on supervised learning from a computational or human ``teacher'' have gained prominence in recent years due to their ability to optimize complex policies for autonomous flight \cite{rmswd-lmruc-13}, video game playing \cite{rgb-rilsp-11,gsllw-amcts-14}, and bipedal locomotion \cite{mlapt-icdcc-15}. Among these methods, guided policy search algorithms \cite{levine2015end} are particularly appealing due to their ability to adapt the teacher to produce data that is best suited for training the final policy with supervised learning. Such algorithms have been used to train complex deep neural network policies for vision-based robotic manipulation \cite{levine2015end}, as well as a variety of other tasks \cite{zkla-ldcpa-15,mlapt-icdcc-15}. However, convergence results for these methods typically follow by construction from their formulation as a constrained optimization, where the teacher is gradually constrained to match the learned policy, and guarantees on the performance of the final policy only hold at convergence if the constraint is enforced exactly. This is problematic in practical applications, where such algorithms are typically executed for a small number of iterations.

In this paper, we show that guided policy search algorithms can be interpreted as approximate variants of mirror descent under constraints imposed by the policy parameterization, with supervised learning corresponding to a projection onto the constraint manifold. Based on this interpretation, we can derive a new, simplified variant of guided policy search, which corresponds exactly to mirror descent under linear dynamics and convex policy spaces. When these convexity and linearity assumptions do not hold, we can show that the projection step is approximate, up to a bound that depends on the step size of the algorithm, which suggests that for a small enough step size, we can achieve continuous improvement. The form of this bound provides us with intuition about how to adjust the step size in practice, so as to obtain a simple algorithm with a small number of hyperparameters.

The main contribution of this paper is a simple new guided policy search algorithm that can train complex, high-dimensional policies by alternating between trajectory-centric reinforcement learning and supervised learning, as well as a connection between guided policy search methods and mirror descent. We also extend previous work on bounding policy cost in terms of KL divergence \cite{rgb-rilsp-11,slmja-trpo-15} to derive a bound on the cost of the policy at each iteration, which provides guidance on how to adjust the step size of the method. We provide empirical results on several simulated robotic navigation and manipulation tasks that show that our method is stable and achieves similar or better performance when compared to prior guided policy search methods, with a simpler formulation and fewer hyperparameters.

\section{Guided Policy Search Algorithms}
\label{sec:gps}



We first review guided policy search methods and background. Policy search algorithms aim to optimize a parameterized policy $\policy_\params(\at|\st)$ over actions $\at$ conditioned on the state $\st$. Given stochastic dynamics $p(\state_{t+1}|\st,\at)$ and cost $\cost(\st,\at)$, the goal is to minimize the expected cost under the policy's trajectory distribution, given by \mbox{$J(\params) = \sum_{t=1}^T E_{\policy_\params(\st,\at)} [\cost(\st,\at)]$}, where we overload notation to use $\policy_\params(\st,\at)$ to denote the marginals of $\policy_\params(\traj) = p(\state_1)\prod_{t=1}^T p(\state_{t+1}|\st,\at) \policy_\params(\at|\st)$, where $\traj =\{\state_1,\action_1,\dots,\state_T,\action_T\}$ denotes a trajectory. A standard reinforcement learning (RL) approach to policy search is to compute the gradient $\nabla_\theta J(\theta)$ and use it to improve $J(\theta)$ \cite{w-ssgfa-92,ps-rlmsp-08}. The gradient is typically estimated using samples obtained from the real physical system being controlled, and recent work has shown that such methods can be applied to very complex, high-dimensional policies such as deep neural networks \cite{slmja-trpo-15,lhphe-ccdrl-16}. However, for complex, high-dimensional policies, such methods tend to be inefficient, and practical real-world applications of such model-free policy search techniques are typically limited to policies with about one hundred parameters \cite{dnp-spsr-13}.

Instead of directly optimizing $J(\theta)$, guided policy search algorithms split the optimization into a ``control phase'' (which we'll call the C-step) that finds multiple simple local policies $\trajdist_i(\at|\st)$ that can solve the task from different initial states $\state_1^i \sim p(\state_1)$, and a ``supervised phase'' (S-step) that optimizes the global policy $\policy_\params(\at|\st)$ to match all of these local policies using standard supervised learning. In fact, a variational formulation of guided policy search \cite{lk-vpsto-13} corresponds to the EM algorithm, where the C-step is actually the E-step, and the S-step is the M-step. The benefit of this approach is that the local policies $\trajdist_i(\at|\st)$ can be optimized separately using domain-specific local methods. Trajectory optimization might be used when the dynamics are known \cite{zkla-ldcpa-15,mlapt-icdcc-15}, while local RL methods might be used with unknown dynamics \cite{la-lnnpg-14,levine2015end}, which still requires samples from the real system, though substantially fewer than the direct approach, due to the simplicity of the local policies. This sample efficiency is the main advantage of guided policy search, which can train policies with nearly a hundred thousand parameters for vision-based control using under 200 episodes \cite{levine2015end}, in contrast to direct deep RL methods that might require orders of magnitude more experience \cite{slmja-trpo-15,lhphe-ccdrl-16}.

\begin{algorithm}[t]
	\caption{Generic guided policy search method \label{alg:gps}}
	\begin{algorithmic}[1]
		\FOR{iteration $k \in \{1, \dots, K\}$}
		\STATE C-step: improve each $\trajdist_i(\at|\st)$ based on surrogate cost $\tilde{\cost}_i(\st,\at)$, return samples $\mathcal{D}_i$ \label{algline:gps_local}
		\STATE S-step: train $\policy_\params(\at|\st)$ with supervised learning on the dataset $\mathcal{D} = \cup_i \mathcal{D}_i$ \label{algline:gps_global}
		\STATE Modify $\tilde{\cost}_i(\st,\at)$ to enforce agreement between $\policy_\params(\at|\st)$ and each $\trajdist(\at|\st)$ \label{algline:gps_dual}
		\ENDFOR
	\end{algorithmic}
\end{algorithm}

A generic guided policy search method is shown in Algorithm~\ref{alg:gps}. The C-step invokes a local policy optimizer (trajectory optimization or local RL) for each $\trajdist_i(\at|\st)$ on line~\ref{algline:gps_local}, and the S-step uses supervised learning to optimize the global policy $\policy_\params(\at|\st)$ on line~\ref{algline:gps_global} using samples from each $\trajdist_i(\at|\st)$, which are generated during the C-step. On line~\ref{algline:gps_dual}, the surrogate cost $\tilde{\cost}_i(\st,\at)$ for each $\trajdist_i(\at|\st)$ is adjusted to ensure convergence. This step is crucial, because supervised learning does not in general guarantee that $\policy_\params(\at|\st)$ will achieve similar long-horizon performance to $\trajdist_i(\at|\st)$ \cite{rgb-rilsp-11}. The local policies might not even be reproducible by a single global policy in general.
To address this issue, most guided policy search methods have some mechanism to force the local policies to agree with the global policy, typically by framing the entire algorithm as a constrained optimization that seeks at convergence to enforce equality between $\policy_\params(\at|\st)$ and each $\trajdist_i(\at|\st)$. The form of the overall optimization problem resembles dual decomposition, and usually looks something like this:
\begin{align}
\min_{\params,\trajdist_1,\dots,\trajdist_N} \sum_{i=1}^N \sum_{t=1}^T E_{\trajdist_i(\st,\at)}[\cost(\st,\at)] \text{ such that } \trajdist_i(\at|\st) = \policy_\params(\at|\st) \,\,\,\forall \st, \at, t, i. \label{eqn:gps}
\end{align}
Since $\state_1^i \sim p(\state_1)$, we have $J(\theta) \approx \sum_{i=1}^N \sum_{t=1}^T E_{\trajdist_i(\st,\at)}[\cost(\st,\at)]$ when the constraints are enforced exactly.
The particular form of the constraint varies depending on the method: prior works have used dual gradient descent \cite{lwa-lnnpg-15}, penalty methods \cite{mlapt-icdcc-15}, ADMM \cite{mt-cbfat-14}, and Bregman ADMM \cite{levine2015end}. We omit the derivation of these prior variants due to space constraints.

\subsection{Efficiently Optimizing Local Policies}

A common and simple choice for the local policies $\trajdist_i(\at|\st)$ is to use time-varying linear-Gaussian controllers of the form $\trajdist_i(\at|\st) = \gauss(\Kpol_t\st + \kpol_t, \ucovart)$, though other options are also possible \cite{mt-cbfat-14,mlapt-icdcc-15,zkla-ldcpa-15}. Linear-Gaussian controllers represent individual trajectories with linear stabilization and Gaussian noise, and are convenient in domains where each local policy can be trained from a different (but consistent) initial state $\state_1^i \sim p(\state_1)$. This represents an additional assumption beyond standard RL, but allows for an extremely efficient and convenient local model-based RL algorithm based on iterative LQR \cite{lt-ilqr-04}. The algorithm proceeds by generating $N$ samples on the real physical system from each local policy $\trajdist_i(\at|\st)$ during the C-step, using these samples to fit local linear-Gaussian dynamics for each local policy of the form $p_i(\state_{t+1}|\st,\at) = \gauss(\fxt\st + \fut\at + \fct,\noise_t)$ using linear regression, and then using these fitted dynamics to improve the linear-Gaussian controller via a modified LQR algorithm \cite{la-lnnpg-14}. This modified LQR method solves the following optimization problem:
\begin{equation}
\min_{\Kpol_t,\kpol_t,\ucovart} \sum_{t=1}^T E_{\trajdist_i(\st,\at)}[\tilde{\cost}_i(\st,\at)] \text{ such that } \kl(\trajdist_i(\traj) \| \bar{\trajdist}_i(\traj)) \leq \epsilon,\label{eqn:lqrkl}
\end{equation}
where we again use $\trajdist_i(\traj)$ to denote the trajectory distribution induced by $\trajdist_i(\at|\st)$ and the fitted dynamics $p_i(\state_{t+1}|\st,\at)$. Here, $\bar{\trajdist}_i(\at|\st)$ denotes the previous local policy, and the constraint ensures that the change in the local policy is bounded, as proposed also in prior works \cite{bs-cps-03,ps-rlmsp-08,pma-reps-10}. This is particularly important when using linearized dynamics fitted to local samples, since these dynamics are not valid outside of a small region around the current controller. In the case of linear-Gaussian dynamics and policies, the KL-divergence constraint $\kl(\trajdist_i(\traj) \| \bar{\trajdist}_i(\traj)) \leq \epsilon$ can be shown to simplify, as shown in prior work~\cite{la-lnnpg-14} and Appendix~\ref{app:kldiv}:
\[
\kl(\trajdist_i(\traj) \| \bar{\trajdist}_i(\traj)) \!=\! \sum_{t=1}^T \kl(\trajdist_i(\at|\st) \| \bar{\trajdist}_i(\at|\st) ) \!=\! \sum_{t=1}^T \!-\! E_{\trajdist_i(\st,\at)} [ \log \bar{\trajdist}_i(\at|\st) ] - \ent(\trajdist_i(\at|\st)),
\]
and the resulting Lagrangian of the problem in Equation~(\ref{eqn:lqrkl}) can be optimized with respect to the primal variables using the standard LQR algorithm, which suggests a simple method for solving the problem in Equation~(\ref{eqn:lqrkl}) using dual gradient descent~\cite{la-lnnpg-14}. The surrogate objective $\tilde{\cost}_i(\st,\at) = \cost(\st,\at) + \phi_i(\theta)$ typically includes some term $\phi_i(\theta)$ that encourages the local policy $\trajdist_i(\at|\st)$ to stay close to the global policy $\policy_\params(\at|\st)$, such as a KL-divergence of the form $\kl(\trajdist_i(\at|\st) \| \policy_\params(\at|\st))$.

\subsection{Prior Convergence Results}

Prior work on guided policy search typically shows convergence by construction, by framing the C-step and S-step as block coordinate ascent on the (augmented) Lagrangian of the problem in Equation~(\ref{eqn:gps}), with the surrogate cost $\tilde{\cost}_i(\st,\at)$ for the local policies corresponding to the (augmented) Lagrangian, and the overall algorithm being an instance of dual gradient descent \cite{lwa-lnnpg-15}, ADMM \cite{mt-cbfat-14}, or Bregman ADMM \cite{levine2015end}. Since these methods enforce the constraint $\trajdist_i(\at|\st) = \policy_\params(\at|\st)$ at convergence (up to linearization or sampling error, depending on the method), we know that $\frac{1}{N}\sum_{i=1}^N E_{\trajdist_i(\st,\at)}[\cost(\st,\at)] \approx E_{\policy_\params(\st,\at)}[\cost(\st,\at)]$ at convergence.\footnote{As mentioned previously, the initial state $\state_1^i$ of each local policy $\trajdist_i(\at|\st)$ is assumed to be drawn from $p(\state_1)$, hence the outer sum corresponds to Monte Carlo integration of the expectation under $p(\state_1)$.} However, prior work does not say anything about $\policy_\params(\at|\st)$ at intermediate iterations, and the constraints of policy search in the real world might often preclude running the method to full convergence. We propose a simplified variant of guided policy search, and present an analysis that sheds light on the performance of both the new algorithm and prior guided policy search methods.


\section{Mirror Descent Guided Policy Search}
\label{sec:md}
\vspace{-0.1in}

\begin{algorithm}[t]
	\caption{Mirror descent guided policy search (MDGPS): convex linear variant \label{alg:mdgps}}
	\begin{algorithmic}[1]
		\FOR{iteration $k \in \{1, \dots, K\}$}
		\STATE C-step: $\trajdist_i \leftarrow \arg\min_{\trajdist_i} E_{\trajdist_i(\traj)}\left[\sum_{t=1}^T \cost(\st,\at)\right] \text{ such that } \kl(\trajdist_i(\traj) \| \policy_\params(\traj)) \leq \epsilon$
		\STATE S-step: $\policy_\params \leftarrow \arg\min_\theta \sum_i \kl(\trajdist_i(\traj) \| \policy_\params(\traj))$ (via supervised learning)
		\ENDFOR
	\end{algorithmic}
\end{algorithm}

In this section, we propose our new simplified guided policy search, which we term mirror descent guided policy search (MDGPS). This algorithm uses the constrained LQR optimization in Equation~(\ref{eqn:lqrkl}) to optimize each of the local policies, but instead of constraining each local policy $\trajdist_i(\at|\st)$ against the previous local policy $\bar{\trajdist}_i(\at|\st)$, we instead constraint it directly against the global policy $\policy_\params(\at|\st)$, and simply set the surrogate cost to be the true cost, such that $\tilde{\cost}_i(\st,\at) = \cost(\st,\at)$. The method is summarized in Algorithm~\ref{alg:mdgps}. In the case of linear dynamics and a quadratic cost (i.e. the LQR setting), and assuming that supervised learning can globally solve a convex optimization problem, we can show that this method corresponds to an instance of mirror descent \cite{bt-mdnps-03} on the objective $J(\theta)$. In this formulation, the optimization is performed on the space of trajectory distributions, with a constraint that the policy must lie on the manifold of policies with the chosen parameterization. Let $\Pi_\Theta$ be the set of all possible policies $\policy_\params$ for a given parameterization, where we overload notation to also let $\Pi_\Theta$ denote the set of trajectory distributions that are possible under the chosen parameterization. The return $J(\theta)$ can be optimized according to $\policy_\params \leftarrow \arg\min_{\policy \in \Pi_\Theta} E_{\policy(\traj)}[\sum_{t=1}^T \cost(\st,\at)]$.
Mirror descent solves this optimization by alternating between two steps at each iteration $k$:
\vspace{-0.05in}
\begin{align*}
\trajdist^k \leftarrow \arg\min_\trajdist E_{\trajdist(\traj)}\left[\sum_{t=1}^T \cost(\st,\at) \right] \text{ s. t. } D\left(\trajdist,\policy^k\right) \leq \epsilon ,\hspace{0.4in}
\policy^{k + 1} \leftarrow \arg\min_{\policy \in \Pi_\Theta} D\left(\trajdist^k, \policy\right).
\end{align*}
The first step finds a new distribution $\trajdist^k$ that minimizes the cost and is close to the previous policy $\policy^k$ in terms of the divergence $D\left(\trajdist,\policy^k\right)$, while the second step projects this distribution onto the constraint set $\Pi_\Theta$, with respect to the divergence $D(\trajdist^k, \policy)$. In the linear-quadratic case with a convex supervised learning phase, this corresponds exactly to Algorithm~\ref{alg:mdgps}: the C-step optimizes $\trajdist^k$, while the S-step is the projection. Monotonic improvement of the global policy $\policy_\params$ follows from the monotonic improvement of mirror descent \cite{bt-mdnps-03}. In the case of linear-Gaussian dynamics and policies, the S-step, which minimizes KL-divergence between trajectory distributions, in fact only requires minimizing the KL-divergence between policies. Using the identity in Appendix~\ref{app:kldiv}, we know that
\begin{equation}
\vspace{-0.05in}
\kl(\trajdist_i(\traj) \| \policy_\params(\traj)) = \sum_{t=1}^T E_{\trajdist_i(\st)}\left[\kl(\trajdist_i(\at|\st) \| \policy_\params(\at|\st))\right]. \label{eqn:mirrorkl}
\end{equation}

\subsection{Implementation for Nonlinear Global Policies and Unknown Dynamics}
\label{sec:implementation}
\vspace{-0.05in}

In practice, we aim to optimize complex policies for nonlinear systems with unknown dynamics. This requires a few practical considerations. The C-step requires a local quadratic cost function, which can be obtained via Taylor expansion, as well as local linear-Gaussian dynamics $p(\state_{t+1}|\st,\at) = \gauss(\fxt\st + \fut\at + \fct,\noise_t)$, which we can fit to samples as in prior work \cite{la-lnnpg-14}. We also need a local time-varying linear-Gaussian approximation to the global policy $\policy_\params(\at|\st)$, denoted $\bar{\policy}_{\params i}(\at|\st)$. This can be obtained either by analytically differentiating the policy, or by using the same linear regression method that we use to estimate $p(\state_{t+1}|\st,\at)$, which is the approach in our implementation. In both cases, we get a different global policy linearization around each local policy. Following prior work \cite{la-lnnpg-14}, we use a Gaussian mixture model prior for both the dynamics and global policy fit.

\begin{algorithm}[t]
	\caption{Mirror descent guided policy search (MDGPS): unknown nonlinear dynamics \label{alg:mdgpsfull}}
	\begin{algorithmic}[1]
		\FOR{iteration $k \in \{1, \dots, K\}$}
		\STATE Generate samples $\mathcal{D}_i = \{\traj_{i,j}\}$ by running either $\trajdist_i$ or $\policy_{\params i}$
		\STATE Fit linear-Gaussian dynamics $p_i(\state_{t+1}|\st,\at)$ using samples in $\mathcal{D}_i$
		\STATE Fit linearized global policy $\bar{\policy}_{\params i}(\at|\st)$ using samples in $\mathcal{D}_i$
		\STATE C-step: $\trajdist_i \leftarrow \arg\min_{\trajdist_i} E_{\trajdist_i(\traj)}[\sum_{t=1}^T \cost(\st,\at)] \text{ such that } \kl(\trajdist_i(\traj) \| \bar{\policy}_{\params i}(\traj)) \leq \epsilon$
		\STATE S-step: $\policy_\params \leftarrow \arg\min_\theta \sum_{t,i,j} \kl(\policy_\params(\at|\state_{t,i,j}) \| \trajdist_i(\at|\state_{t,i,j}))$ (via supervised learning)
		\STATE Adjust $\epsilon$ (see Section~\ref{sec:step})
		\ENDFOR
	\end{algorithmic}
\end{algorithm}

The S-step can be performed approximately in the nonlinear case by using the samples collected for dynamics fitting to also train the global policy. Following prior work \cite{levine2015end}, our S-step minimizes\footnote{Note that we flip the KL-divergence inside the expectation, following~\cite{levine2015end}. We found that this produced better results. The intuition behind this is that, because $\log \trajdist_i(\at|\st)$ is proportional to the Q-function of $\trajdist_i(\at|\st)$ (see Appendix~\ref{app:cstep}), $\kl(\policy_\params(\at|\state_{t,i,j}) \| \trajdist_i(\at|\state_{t,i,j})$ minimizes the cost-to-go under $\trajdist_i(\at|\st)$ with respect to $\policy_\params(\at|\st)$, which provides for a more informative objective than the unweighted likelihood in Equation~(\ref{eqn:mirrorkl}).}
\[
\sum_{i,t} E_{\trajdist_i(\st)}\left[\kl(\policy_\params(\at|\st) \| \trajdist_i(\at|\st)) \right] \approx \frac{1}{|\mathcal{D}_i|}\sum_{i,t,j} \kl(\policy_\params(\at|\state_{t,i,j}) \| \trajdist_i(\at|\state_{t,i,j})),
\]
\noindent where $\state_{t,i,j}$ is the $j^\text{th}$ sample from $\trajdist_i(\st)$ obtained by running $\trajdist_i(\at|\st)$ on the real system. For linear-Gaussian $\trajdist_i(\at|\st)$ and (nonlinear) conditionally Gaussian $\policy_\params(\at|\st) = \gauss(\polmu(\st),\polsig(\st))$, where $\polmu$ and $\polsig$ can be any function (such as a deep neural network), the KL-divergence $\kl(\policy_\params(\at|\state_{t,i,j}) \| \trajdist_i(\at|\state_{t,i,j}))$ can easily be evaluated and differentiated in closed form~\cite{levine2015end}. However, in the nonlinear setting, minimizing this objective no longer minimizes the KL-divergence between trajectory distributions $\kl(\policy_\params(\traj) \| \trajdist_i(\traj))$ exactly, which means that MDGPS does not correspond exactly to mirror descent: although the C-step can still be evaluated exactly, the S-step now corresponds to an approximate projection onto the constraint manifold. In the next section, we discuss how we can bound the error in this projection. A summary of the nonlinear MDGPS method is provided in Algorithm~\ref{alg:mdgpsfull}, and additional details are in Appendix~\ref{app:mdgpsdetails}. The samples for linearizing the dynamics and policy can be obtained by running either the last local policy $\trajdist_i(\at|\st)$, or the last global policy $\policy_\params(\at|\st)$. Both variants produce good results, and we compare them in Section~\ref{sec:experiments}.

\subsection{Analysis of Prior Guided Policy Search Methods as Approximate Mirror Descent}

The main distinction between the proposed method and prior guided policy search methods is that the constraint $\kl(\trajdist_i(\traj) \| \bar{\policy}_{\params i}(\traj)) \leq \epsilon$ is enforced on the local policies at each iteration, while in prior methods, this constraint is iteratively enforced via a dual descent procedure over multiple iterations. This means that the prior methods perform approximate mirror descent with step sizes that are adapted (by adjusting the Lagrange multipliers) but not constrained exactly. In our empirical evaluation, we show that our approach is somewhat more stable, though sometimes slower than these prior methods. This empirical observation agrees with our intuition: prior methods can sometimes be faster, because they do not exactly constrain the step size, but our method is simpler, requires less tuning, and always takes bounded steps on the global policy in trajectory space.

\section{Analysis in the Nonlinear Case}

Although the S-step under nonlinear dynamics is not an optimal projection onto the constraint manifold, we can bound the additional cost incurred by this projection in terms of the KL-divergence between $\trajdist_i(\at|\st)$ and $\policy_\params(\at|\st)$. This analysis also reveals why prior guided policy search algorithms, which only have asymptotic convergence guarantees, still attain good performance in practice even after a small number of iterations. We will drop the subscript $i$ from $\trajdist_i(\at|\st)$ in this section for conciseness, though the same analysis can be repeated for multiple local policies $\trajdist_i(\at|\st)$.

\subsection{Bounding the Global Policy Cost}
\label{sec:bounding}

The analysis in this section is based on the following lemma, which we prove in Appendix~\ref{app:distro_bound}, building off of earlier results by Ross et al.~\cite{rgb-rilsp-11} and Schulman et al.~\cite{slmja-trpo-15}:

\begin{lemma}
	\label{lemma:distro_bound}
	Let $\epsilon_t = \max_{\st} \kl(\trajdist(\at|\st) \| \policy_\params(\at|\st)$. Then $\tv(\trajdist(\st)\| \policy_\params(\st)) \leq 2 \sum_{t=1}^T \sqrt{2\epsilon_t}$.
\end{lemma}


This means that if we can bound the KL-divergence between the policies, then the total variation divergence between their state marginals (given by $\tv(\trajdist(\st)\| \policy_\params(\st)) = \frac{1}{2}\|\trajdist(\st) - \policy_\params(\st)\|_1$) will also be bounded. This bound allows us in turn to relate the total expected costs of the two policies to each other according to the following lemma, which we prove in Appendix~\ref{app:cost_bound}:

\begin{lemma}
	\label{lemma:cost_bound}
	If $\tv(\trajdist(\st)\| \policy_\params(\st)) \leq 2 \sum_{t=1}^T \sqrt{2\epsilon_t}$, then we can bound the total cost of $\policy_\params$ as
    \[
    \sum_{t=1}^T E_{\policy_\params(\st,\at)}[\cost(\st,\at)] \leq \sum_{t=1}^T \left[ E_{p(\st,\at)}[\cost(\st,\at)] + \sqrt{2\epsilon_t} \max_{\st, \at} \cost(\st, \at) + 2\sqrt{2\epsilon_t} Q_{\text{max},t} \right]
    \]
	where $Q_{\text{max},t} = \sum_{t^\prime=t}^T \max_{\state_{t^\prime},\action_{t^\prime}}\cost(\state_{t^\prime},\action_{t^\prime})$, the maximum total cost from time $t$ to $T$.
\end{lemma}

This bound on the cost of $\policy_\params(\at|\st)$ tells us that if we update $\trajdist(\at|\st)$ so as to decrease its total cost or decrease its KL-divergence against $\policy_\params(\at|\st)$, we will eventually reduce the cost of $\policy_\params(\at|\st)$. For the MDGPS algorithm, this bound suggests that we can ensure improvement of the global policy within a small number of iterations by appropriately choosing the constraint $\epsilon$ during the C-step. Recall that the C-step constrains $\sum_{t=1}^T \epsilon_t \leq \epsilon$, so if we choose $\epsilon$ to be small enough, we can close the gap between the local and global policies. Optimizing the bound directly turns out to produce very slow learning in practice, because the bound is very loose. However, it tells us that we can either decrease $\epsilon$ toward the end of the optimization process or if we observe the global policy performing much worse than the local policies. We discuss how this idea can be put into action in the next section.


\subsection{Step Size Selection}
\label{sec:step}

In prior work \cite{lwa-lnnpg-15}, the step size $\epsilon$ in the local policy optimization is adjusted by considering the difference between the predicted change in the cost of the local policy $\trajdist(\at|\st)$ under the fitted dynamics, and the actual cost obtained when sampling from that policy. The intuition is that, because the linearized dynamics are local, we incur a larger cost the further we deviate from the previous policy. We can adjust the step size by estimating the rate at which the additional cost is incurred and choose the optimal tradeoff. Let $\ell_{k-1}^{k-1}$ denote the expected cost under the previous local policy $\bar{\trajdist}(\at|\st)$, $\ell_{k-1}^k$ the cost under the current local policy $\trajdist(\at|\st)$ and the previous fitted dynamics (which were estimated using samples from $\bar{\trajdist}(\at|\st)$ and used to optimize $\trajdist(\at|\st)$), and $\ell_k^k$ the cost of the current local policy under the dynamics estimated using samples from $\trajdist(\at|\st)$ itself. Each of these can be computed analytically under the linearized dynamics. We can view the difference $\ell_k^k - \ell_{k-1}^k$ as the additional cost we incur from imperfect dynamics estimation. Previous work suggested modeling the change in cost as a function of $\epsilon$ as following: $\ell_k^k - \ell_{k-1}^{k-1} = a\epsilon^2 + b\epsilon$, where $b$ is the change in cost per unit of KL-divergence, and $a$ is additional cost incurred due to inaccurate dynamics \cite{lwa-lnnpg-15}. This model is reasonable because the integral of a quadratic cost under a linear-Gaussian system changes roughly linearly with KL-divergence. The additional cost due to dynamics errors is assumes to scale superlinearly, allowing us to solve for $b$ by looking at the difference $\ell_k^k - \ell_{k-1}^k$ and then solving for a new optimal $\epsilon^\prime$ according to $\epsilon^\prime = -b/2a$, resulting in the update $\epsilon^\prime = \epsilon (\ell_{k-1}^k - \ell_{k-1}^{k-1}) / 2(\ell_{k-1}^k - \ell_k^k)$.

In MDGPS, we propose to use two step size adjustment rules. The first rule simply adapts the previous method to the case where we constrain the new local policy $\trajdist(\at|\st)$ against the global policy $\policy_\params(\at|\st)$, instead of the previous local policy $\bar{\trajdist}(\at|\st)$. In this case, we simply replace $\ell_{k-1}^{k-1}$ with the expected cost under the previous global policy, given by $\ell_{k-1}^{k-1,\policy}$, obtained using its linearization $\bar{\policy}_\params(\at|\st)$. We call this the ``classic'' step size: $\epsilon^\prime = \epsilon (\ell_{k-1}^{k} - \ell_{k-1}^{k-1,\policy}) / 2(\ell_{k-1}^k - \ell_k^k)$.

However, we can also incorporate intuition from the bound in the previous section to obtain a more conservative step adjustment that reduces $\epsilon$ not only when the obtained local policy improvement doesn't meet expectations, but also when we detect that the global policy is unable to reproduce the behavior of the local policy. In this case, reducing $\epsilon$ reduces the KL-divergence between the global and local policies which, as shown in the previous section, tightens the bound on the global policy return. As mentioned previously, directly optimizing the bound tends to perform poorly because the bound is quite loose. However, if we estimate the cost of the global policy using its linearization, we can instead adjust the step size based on a simple model of \emph{global} policy cost. We use the same model for the change in cost, given by $\ell_k^{k,\policy} - \ell_{k-1}^{k-1,\policy} = a\epsilon^2 + b\epsilon$. However, for the term $\ell_k^k$, which reflects the actual cost of the new policy, we instead use the cost of the new global policy $\ell_k^{k,\policy}$, so that $a$ now models the additional loss due to \emph{both} inaccurate dynamics and inaccurate projection: if $\ell_k^{k,\policy}$ is much worse than $\ell_{k-1}^{k}$, then either the dynamics were too local, or S-step failed to match the performance of the local policies. In either case, we decrease the step size.\footnote{Although we showed before that the discrepancy depends on $\sum_{t=1}^T \sqrt{2\epsilon}_t$, here we use $\epsilon^2$. This is a simplification, but the net result is the same: when the global policy is worse than expected, $\epsilon$ is reduced.} As before, we can solve for the new step size $\epsilon^\prime$ according to $\epsilon^\prime = \epsilon (\ell_{k-1}^{k} - \ell_{k-1}^{k-1,\policy}) / 2(\ell_{k-1}^{k} - \ell_k^{k,\policy})$. We call this the ``global'' step size. Details of how each quantity in this equation is computed are provided in Appendix~\ref{app:step}.

\section{Relation to Prior Work}
\label{sec:prior_work}

While we've discussed the connections between MDGPS and prior guided policy search methods, in this section we'll also discuss the connections between our method and other policy search methods. One popular supervised policy learning methods is DAGGER~\cite{rgb-rilsp-11}, which also trains the policy using supervised learning, but does not attempt to adapt the teacher to provide better training data. MDGPS removes the assumption in DAGGER that the supervised learning stage has bounded error against an arbitrary teacher policy. MDGPS does not need to make this assumption, since the teacher can be adapted to the limitations of the global policy learning. This is particularly important when the global policy has computational or observational limitations, such as when learning to use camera images for partially observed control tasks or, as shown in our evaluation, blind peg insertion.

When we sample from the global policy $\policy_\params(\at|\st)$, our method resembles policy gradient methods with KL-divergence constraints \cite{ps-rlmsp-08,pma-reps-10,slmja-trpo-15}. However, policy gradient methods update the policy $\policy_\params(\at|\st)$ at each iteration by linearizing with respect to the policy parameters, which often requires small steps for complex, nonlinear policies, such as neural networks. In contrast, we linearize in the space of time-varying linear dynamics, while the policy is optimized at each iteration with many steps of supervised learning (e.g. stochastic gradient descent). This makes MDGPS much better suited for quickly and efficiently training highly nonlinear, high-dimensional policies.





\section{Experimental Evaluation}
\label{sec:experiments}


We compare several variants of MDGPS and a prior guided policy search method based on Bregman ADMM (BADMM) \cite{levine2015end}. We evaluate all methods on one simulated robotic navigation task and two manipulation tasks. Guided policy search code, including BADMM and MDGPS methods, is available at \texttt{https://www.github.com/cbfinn/gps}.

\begin{wrapfigure}{r}{0.25\textwidth}
	\vspace{-0.18in}
    \includegraphics[width=0.24\textwidth]{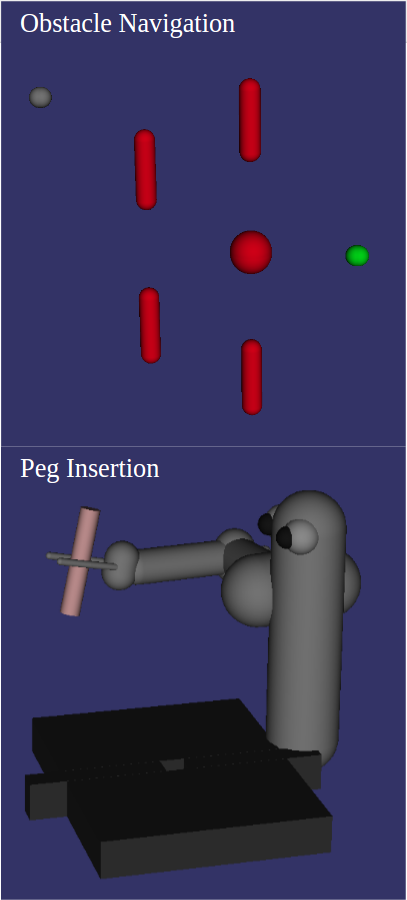}
    \vspace{-0.30in}
\end{wrapfigure}

\vspace{-0.12in}
\paragraph{Obstacle Navigation.} In this task, a 2D point mass (grey) must navigate around obstacles to reach a target (shown in green), using velocities and positions relative to the target. We use $N=5$ initial states, with 5 samples per initial state per iteration. The target and obstacles are fixed, but the starting position varies.

\vspace{-0.12in}
\paragraph{Peg Insertion.} This task, which is more complex, requires controlling a 7 DoF 3D arm to insert a tight-fitting peg into a hole. The hole can be in different positions, and the state consists of joint angles, velocities, and end-effector positions relative to the target. This task is substantially more challenging physically. We use $N=9$ different hole positions, with 5 samples per initial state per iteration.

\vspace{-0.12in}
\paragraph{Blind Peg Insertion.} The last task is a blind variant of the peg insertion task, where the target-relative end effector positions are provided to the local policies, but not to the global policy $\policy_\params(\at|\st)$. This requires the global policy to search for the hole, since no input to the global policy can distinguish between the different initial state $\state_1^i$. This makes it much more challenging to adapt the global and local policies to each other, and makes it impossible for the global learner to succeed without adaptation of the local policies. We use $N=4$ different hole positions, with 5 samples per initial state per iteration.

\begin{figure}
    \centering
        \includegraphics[width=\textwidth]{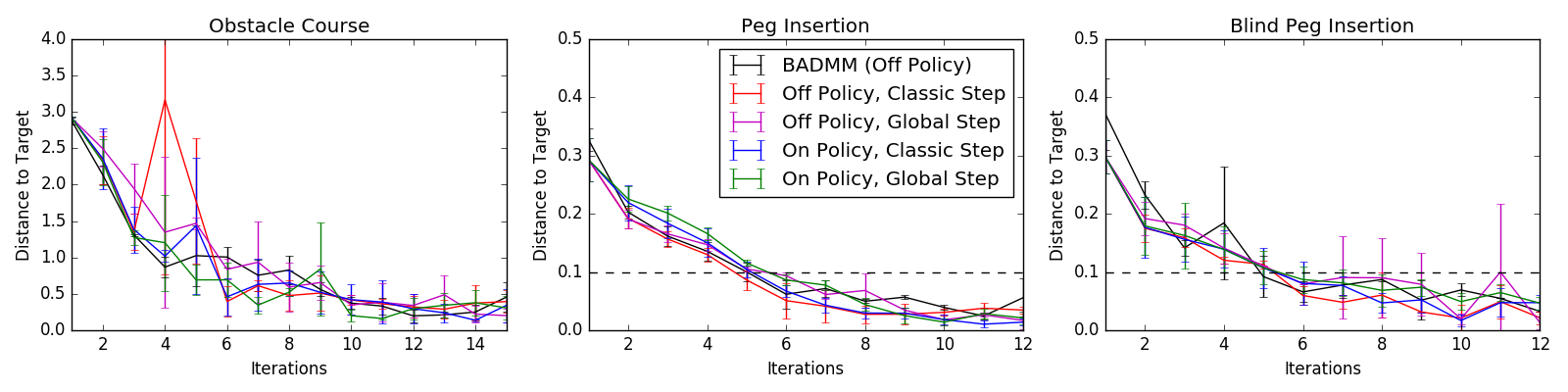}
    \caption{Results for MDGPS variants and BADMM on each task. MDGPS is tested with local policy (``off policy'') and global policy (``on policy'') sampling (see Section~\ref{sec:implementation}), and both the ``classic'' and ``global'' step sizes (see Section~\ref{sec:step}). The vertical axis for the obstacle task shows the average distance between the point mass and the target. The vertical axis for the peg tasks shows the average distance between the bottom of the peg and the hole. Distances above 0.1, which is the depth of the hole (shown as a dotted line) indicate failure. All experiments are repeated three times, with the average performance and standard deviation shown in the plots.}
    \label{fig:results}
    \vspace{-0.1in}
\end{figure}

The global policy for each task consists of a fully connected neural network with two hidden layers with $40$ rectified linear units. The same settings are used for MDGPS and the prior BADMM-based method, except for the difference in surrogate costs, constraints, and step size adjustment methods discussed in the paper. Results are presented in Figure~\ref{fig:results}. On the easier point mass and peg tasks, all of the methods achieve similar performance. However, the MDGPS methods are all substantially easier to apply to these tasks, since they have very few free hyperparameters. An initial step size must be selected, but the adaptive step size adjustment rules make this choice less important. In contrast, the BADMM method requires choosing an initial weight on the augmented Lagrangian term, an adjustment schedule for this term, a step size on the dual variables, and a step size for local policies, all of which have a substantial impact on the final performance of the method (the reported results are for the best setting of these parameters, identified with a hyperparameter sweep).

On the harder blind peg task, MDGPS consistently outperforms BADMM when sampling from the local policies (``off policy''), with both the classic and global step sizes. This is particularly apparent in the success rates in Table~\ref{tbl:success}, which shows that the MDGPS policies succeed at actually inserting the peg into the hole more often and on more conditions. This suggests that our method is better able to improve global policies particularly in situations where informational or representational constraints make na\"{i}ve imitation of the local policies insufficient to solve the task. On-policy sampling tends to learn slower, since the approximate projection causes the global policy to lag behind the local policy in performance, but this method is still able to consistently improve the global policies. Sampling from the global policies may be desirable in practice, since the global policies can directly use observations at runtime instead of requiring access to the state \cite{levine2015end}. The global step size also tends to be more conservative, but produces more consistent and monotonic improvement.




\begin{table}

{\footnotesize
\begin{tabular}{| c | c | c | c | c | c | c |}
\hline
& Iteration &BADMM & Off Pol., Classic & Off Pol., Global & On Pol., Classic & On Pol., Global\\
\hline
\multirow{4}{*}{\rotatebox{90}{\scriptsize{Peg}}}
& 3 & 0.00~\% & 0.00~\% & 0.00~\% & 0.00~\% & 0.00~\% \\
& 6 & 51.85~\% & \textbf{62.96~\%} & 22.22~\% & 48.15~\% & 33.33~\% \\
& 9 & 51.85~\% & 77.78~\% & 74.07~\% & 77.78~\% & \textbf{81.48~\%} \\
& 12 & 77.78~\% & 70.73~\% & \textbf{92.59~\%} & \textbf{92.59~\%} & 85.19~\% \\
\hline
\hline

\multirow{4}{*}{\rotatebox{90}{\scriptsize{Blind Peg}}}
& 3& 0.00~\% & 0.00~\% & 0.00~\% & 0.00~\% & 0.00~\% \\
& 6& 50.00~\% & \textbf{58.33~\%} & 25.00~\% & 33.33~\% & 25.00~\% \\
& 9& 58.33~\% & \textbf{75.00~\%} & 50.00~\% & 58.33~\% & 33.33~\% \\
& 12& 75.00~\% & 83.33~\% & \textbf{91.67~\%} & 58.33~\% & 58.33~\% \\
\hline
\end{tabular}
}

\vspace{0.1in}
\caption{
Success rates of each method on each peg insertion task. Success is defined as inserting the peg into the hole with a final distance of less than 0.06. Results are averaged over three runs.}
\label{tbl:success}	

\vspace{-0.3in}
\end{table}

\section{Discussion and Future Work}
\label{sec:discussion}

We presented a new guided policy search method that corresponds to mirror descent under linearity and convexity assumptions, and showed how prior guided policy search methods can be seen as approximating mirror descent. We provide a bound on the return of the global policy in the nonlinear case, and argue that an appropriate step size can provide improvement of the global policy in this case also. Our analysis provides us with the intuition to design an automated step size adjustment rule, and we illustrate empirically that our method achieves good results on a complex simulated robotic manipulation task while requiring substantially less tuning and hyperparameter optimization than prior guided policy search methods. Manual tuning and hyperparameter searches are a major challenge across a range of deep reinforcement learning algorithms, and developing scalable policy search methods that are simple and reliable is vital to enable further progress.

As discussed in Section~\ref{sec:prior_work}, MDGPS has interesting connections to other policy search methods. Like DAGGER \cite{rgb-rilsp-11}, MDGPS uses supervised learning to train the policy, but unlike DAGGER, MDGPS does not assume that the learner is able to reproduce an arbitrary teacher's behavior with bounded error, which makes it very appealing for tasks with partial observability or other limits on information, such as learning to use camera images for robotic manipulation \cite{levine2015end}. When sampling directly from the global policy, MDGPS also has close connections to policy gradient methods that take steps of fixed KL-divergence \cite{ps-rlmsp-08,slmja-trpo-15}, but with the steps taken in the space of trajectories rather than policy parameters, followed by a projection step. In future work, it would be interesting to explore this connection further, so as to develop new model-free policy gradient methods.



\bibliographystyle{plain}
\bibliography{references}

\appendix

\section{KL Divergence Between Gaussian Trajectory Distributions}
\label{app:kldiv}

In this appendix, we derive the KL-divergence between two Gaussian trajectory distributions corresponding to time-varying linear-Gaussian dynamics $p(\state_{t+1}|\st,\at)$ and two policies $p(\at|\st)$ and $q(\at|\st)$. The two policies induce Gaussian trajectory distributions (with block-diagonal covariances) according to
\[
p(\traj) = p(\state_1)\prod_{t=1}^T p(\state_{t+1}|\st,\at)p(\at|\st), \hspace{0.2in} q(\traj) = p(\state_1)\prod_{t=1}^T p(\state_{t+1}|\st,\at)q(\at|\st).
\]
We can therefore derive their KL-divergence as
\begin{align*}
\kl(p(\traj) \| q(\traj)) &= E_{p(\traj)}\left[\log p(\traj) - \log q(\traj)\right] \\
&= E_{p(\traj)}\left[\sum_{t=1}^T \log p(\at|\st) - \log q(\at|\st)\right] \\
&= \sum_{t=1}^T E_{p(\st,\at)} \left[ \log p(\at|\st) - \log q(\at|\st) \right] \\
&= \sum_{t=1}^T - E_{p(\st,\at)} \left[ \log q(\at|\st) \right] - E_{p(\st)}[\ent(p(\at|\st))]\\
&= \sum_{t=1}^T - E_{p(\st,\at)} \left[ \log q(\at|\st) \right] - \ent(p(\at|\st))
\end{align*}
where the second step follows because the dynamics and initial state distribution cancel, the third step follows by linearity of expectations, the fourth step from the definition of differential entropy, and the last step follows from the fact that the entropy of a conditional Gaussian distribution is independent on the quantity that it is conditioned on, since it depends only on the covariance and not the mean. We therefore have
\[
\kl(p(\traj) \| q(\traj)) = \sum_{t=1}^T - E_{p(\st,\at)} \left[ \log q(\at|\st) \right] - \ent(p(\at|\st)).
\]
By the definition of KL-divergence, we can also write this as
\[
\kl(p(\traj) \| q(\traj)) = \sum_{t=1}^T E_{p(\st,\at)} \left[ \kl(p(\at|\st) \| q(\at \| \st)) \right].
\]


\section{Details of the MDGPS Algorithm}
\label{app:mdgpsdetails}

A summary of the MDGPS algorithm appears in Algorithm~\ref{alg:mdgps}, and is repeated below for convenience:

\begin{algorithm}[H]
	\caption{Mirror descent guided policy search (MDGPS): unknown nonlinear dynamics \label{alg:mdgpsfull}}
	\begin{algorithmic}[1]
		\FOR{iteration $k \in \{1, \dots, K\}$}
		\STATE Generate samples $\mathcal{D}_i = \{\traj_{i,j}\}$ by running either $\trajdist_i$ or $\policy_{\params i}$
		\STATE Fit linear-Gaussian dynamics $p_i(\state_{t+1}|\st,\at)$ using samples in $\mathcal{D}_i$
		\STATE Fit linearized global policy $\bar{\policy}_{\params i}(\at|\st)$ using samples in $\mathcal{D}_i$
		\STATE C-step: $\trajdist_i \leftarrow \arg\min_{\trajdist_i} E_{\trajdist_i(\traj)}[\sum_{t=1}^T \cost(\st,\at)] \text{ such that } \kl(\trajdist_i(\traj) \| \bar{\policy}_{\params i}(\traj)) \leq \epsilon$
		\STATE S-step: $\policy_\params \leftarrow \arg\min_\theta \sum_{t,i,j} \kl(\policy_\params(\at|\state_{t,i,j}) \| \trajdist_i(\at|\state_{t,i,j}))$ (via supervised learning)
		\STATE Adjust $\epsilon$ (see Section~\ref{sec:step})
		\ENDFOR
	\end{algorithmic}
\end{algorithm}

\subsection{C-Step Details}
\label{app:cstep}

The C-step solves the following constrained optimization problem:
\[
\trajdist_i \leftarrow \arg\min_{\trajdist_i} E_{\trajdist_i(\traj)}\left[\sum_{t=1}^T \cost(\st,\at)\right] \text{ such that } \kl(\trajdist_i(\traj) \| \bar{\policy}_{\params i}(\traj)) \leq \epsilon.
\]
The solution to this problem follows prior work~\cite{la-lnnpg-14}, and is reviewed here for completeness. First, the Lagrangian of this problem is given by
\begin{align*}
\obj(\trajdist_i,\eta) &= E_{\trajdist_i(\traj)}\left[\sum_{t=1}^T \cost(\st,\at)\right] + \eta( \kl(\trajdist_i(\traj) \| \bar{\policy}_{\params i}(\traj)) - \epsilon) \\
&= \sum_{t=1}^T E_{\trajdist_i(\st,\at)} [ \cost(\st,\at) - \eta \log \bar{\policy}_{\params i}(\at|\st) ] - \eta\ent(\trajdist(\at|\st)) - \eta\epsilon,
\end{align*}
where equality follows from the identity in Appendix~\ref{app:kldiv}. As discussed in prior work~\cite{la-lnnpg-14}, we can minimize this Lagrangian with respect to $\trajdist_i$ by solving an LQR problem (assuming a quadratic expansion of $\cost(\st,\at)$) with a surrogate cost
\[
\tilde{\cost}(\st,\at) = \frac{1}{\eta}\cost(\st,\at) - \log\bar{\policy}_{\params i}(\at|\st).
\]
This follows because LQR can be shown to solve the following problem~\cite{la-lnnpg-14}
\[
\trajdist_ i = \arg\min_{\trajdist_i} \sum_{t=1}^T E_{\trajdist_i(\st,\at)} \left[ \tilde{\cost}(\st,\at) \right] - \ent(\trajdist_i(\at|\st))
\]
if we set $\trajdist_i(\at|\st) = \gauss(\Kpol_t \st + \kpol_t, \Quut\inv)$, where $\Kpol_t$ and $\kpol_t$ are the optimal feedback and feedforward terms, respectively, and $\Quut$ is the action component of the Q-function matrix computed by LQR, where the full Q-function is given by
\[
Q(\st,\at) = \frac{1}{2}\st\tr\Qxxt\st + \frac{1}{2}\at\tr\Quut\at + \at\tr\Quxt\st + \st\tr\Qxt + \at\tr\Qut.
\]
This maximum entropy LQR solution also directly from the so-called Kalman duality, which describes a connection between LQR and Kalman smoothing.

Once we can minimize the Lagrangian with respect to $\trajdist_i$, we can solve the original constrained problem by using dual gradient descent to iteratively adjust the dual variable $\eta$. Since there is only a single dual variable, we can find it very efficiently by using a bracketing line search, exploiting the fact that the dual function is convex.

As discussed in the paper, the dynamics $\trajdist_i(\state_{t+1}|\st,\at)$ are estimated by using samples (drawn from either the local policy or the global policy) and linear regression. Following prior work~\cite{la-lnnpg-14}, the dynamics at each step are fitted using linear regression with a Gaussian mixture model prior. This prior incorporates samples from other time steps and previous iterations to allow the regression procedure to use a very small number of sampled trajectories.

\subsection{S-Step Details}

The step solves the following optimization problem:
\[
\policy_\params \leftarrow \arg\min_\theta \sum_{t,i,j} \kl(\policy_\params(\at|\state_{t,i,j}) \| \trajdist_i(\at|\state_{t,i,j})).
\]
Since both $\policy_\params(\at|\st) = \gauss(\polmu(\st),\polsig(\st))$ and $\trajdist_i(\at|\st) = \gauss(\Kpol_{ti}\st + \kpol_{ti}, \ucovar_{ti})$ are assumed to be conditionally Gaussian, this objective can be rewritten in closed form as
\begin{align*}
\policy_\params \leftarrow \arg\min_\theta \sum_{t,i,j}\,\,\, &\trace[\ucovar_{ti}\inv\polsig(\state_{t,i,j})] - \log|\polsig(\state_{t,i,j})| + \\
&(\polmu(\state_{t,i,j}) - \trajmu_{ti}(\state_{t,i,j}))\ucovar_{ti}\inv(\polmu(\state_{t,i,j}) - \trajmu_{ti}(\state_{t,i,j})).
\end{align*}
Note that the last term is simply a weighted quadratic cost on the policy mean $\polmu(\state_{t,i,j}$, which lends itself to simple and straightforward optimization using stochastic gradient descent. In our implementation, we use a policy where the covariance $\polsig(\st)$ is independent of the state $\st$, and therefore we can solve for the covariance in closed form, as discussed in prior work~\cite{levine2015end}. However, in general, the covariance could also be optimized using stochastic gradient descent.

\subsection{Step Size Adjustment}
\label{app:step}

As discussed in Section~\ref{sec:step}, the step size adjustment procedure requires estimating quantities of the type $\ell_m^k = \sum_{t=1}^T E_{\trajdist^k(\st,\at)}[\cost(\st,\at)]$, where $\trajdist^k(\st,\at)$ is the marginal of the local policy used to generate samples at iteration $k$ and the dynamics fitted at iteration $m$ (not to be confused with $\trajdist_i$, which we use to denote the local policy for the $i^\text{th}$ initial state, independent of the iteration number). We also use terms of the form $\ell_m^{k,\policy} = \sum_{t=1}^T E_{\bar{\policy}_\params^k(\st,\at)}[\cost(\st,\at)]$, which give the expected cost under the dynamics at iteration $m$ and the linearized global policy at iteration $k$. Specifically, we require $\ell_{k-1}^{k-1}$, $\ell_{k-1}^k$, and $\ell_k^k$, as well as the corresponding global policy terms $\ell_{k-1}^{k-1,\policy}$, $\ell_{k-1}^{k,\policy}$, and $\ell_k^{k,\policy}$.

All of these terms can be computed analytically, since the fitted dynamics, local policies, and linearized global policy $\bar{\policy}_\params(\at|\st)$ are all linear-Gaussian. The state-action marginals $\trajdist(\st,\at)$ in linear-Gaussian policies can be computed simply by propagating Gaussian densities forward in time, according to
\begin{align*}
\mu_{\st,\at} &= \left[ \begin{array}{l} \mu_{\st} \\ \Kpol_t \mu_{\st} + \kpol_t \end{array} \right] & \Sigma_{\st,\at} &= \left[ \begin{array}{l l}
\Sigma_{\st} & \Sigma_{\st}\Kpol_t\tr \\
\Kpol_t \Sigma_{\st} & \Kpol_t \Sigma_{\st} \Kpol_t\tr + \ucovar_t
\end{array}
\right] \\
\mu_{\state_{t+1}} &= \ft \mu_{\st,\at} + \fct & \Sigma_{\state_{t+1}} &= \ft \Sigma_{\st,\at} \ft\tr + \noise_t
\end{align*}
where we have $p(\state_{t+1}|\st,\at) = \gauss(\ft(\st,\at)\tr + \fct,\noise_t)$ and $\trajdist(\at|\st) = \gauss(\Kpol_t\st + \kpol_t,\ucovart)$, and then we can estimate the expectation of the cost at time $t$ simply by integrating the quadratic cost under the Gaussian state-action marginals.

\section{Global Policy Cost Bounds}
\label{app:bounds}

In this appendix, we prove the bound on the policy cost discussed in Section~\ref{sec:bounding}. The proof combines the earlier results from Ross et al.~\cite{rgb-rilsp-11} and Schulman et al.~\cite{slmja-trpo-15}, and extends them to the case of finite-horizon episodic tasks.

\subsection{Policy State Distribution Bound}
\label{app:distro_bound}

We begin by proving Lemma~\ref{lemma:distro_bound}, which we restate below with slightly simplified notation, replacing $\policy_\params$ by $q$:

\begin{lemma}
	Let $\epsilon_t = \max_{\st} \kl(p(\at|\st) \| q(\at|\st)$. Then $\tv(p(\st)\| q(\st)) \leq 2 \sum_{t=1}^T \sqrt{2\epsilon_t}$.
\end{lemma}

The proof first requires introducing a lemma that relates the total variation divergence $\max_{\st} \|p(\at|\st) - q(\at|\st)\|_1$ between two policies to the probability that the policies will take the same action in a discrete setting (extensions to the continuous setting are also possible):

\begin{lemma}
	\label{lemma:mix}
	Assume that $\max_{\st} \|p(\at|\st) - q(\at|\st)\|_1 \leq \sqrt{2\epsilon_t}$, then the probability that $p$ and $q$ take the same action at time step $t$ is $1-\sqrt{2\epsilon_t}$.
\end{lemma}

The proof for this lemma was presented by Schulman et al.~\cite{slmja-trpo-15}. We can use it to bound the state distribution difference as following. First, we are acting according to $p(\at|\st)$, the probability that the same action would have been taken by $q(\at|\st)$, based on Lemma~\ref{lemma:mix}, is $(1-\sqrt{2\epsilon_t})$, so the probability that all actions up to time $t$ would have been taken by $q(\at|\st)$ is given by $\prod_{t^\prime = 1}^t (1 - \sqrt{2\epsilon_{t^\prime}})$. We can therefore express the state distribution $p(\st)$ as
\begin{align*}
p(\st) &= \left[\prod_{t^\prime = 1}^t (1 - \sqrt{2\epsilon_{t^\prime}})\right] q(\st) + \left(1 - \prod_{t^\prime = 1}^t (1 - \sqrt{2\epsilon_{t^\prime}}) \right) \tilde{p}(\st) \\
&= \left[\prod_{t^\prime = 1}^t (1 - \sqrt{2\epsilon_{t^\prime}})\right] [q(\st) - \tilde{p}(\st)] + \tilde{p}(\st),
\end{align*}
\noindent where $\tilde{p}(\st)$ is some other distribution. In order to bound $\tv(p(\st) \| q(\st)) = \|p(\st) - q(\st)\|_1$, we can substitute this equation into $\|p(\st) - q(\st)\|_1$ to get
\begin{align*}
\|p(\st) - q(\st)\|_1 &= \left\| \left[\prod_{t^\prime = 1}^t (1 - \sqrt{2\epsilon_{t^\prime}})\right] [q(\st) - \tilde{p}(\st)] + \tilde{p}(\st) - q(\st) \right\| \\
&= \left\| \left[1 - \prod_{t^\prime = 1}^t (1 - \sqrt{2\epsilon_{t^\prime}})\right] [q(\st) - \tilde{p}(\st)] \right\| \\
&= \left[1 - \prod_{t^\prime = 1}^t (1 - \sqrt{2\epsilon_{t^\prime}})\right] \left\| q(\st) - \tilde{p}(\st) \right\| \\
&\leq 2\left[1 - \prod_{t^\prime = 1}^t (1 - \sqrt{2\epsilon_{t^\prime}})\right],
\end{align*}
where the last inequality comes from the fact that $\left\| q(\st) - \tilde{p}(\st) \right\| \leq 2$ for discrete distributions. With continuous densities, we could extend the result by taking the limit of an infinitely fine discretization. Next, we note that
\[
\prod_{t^\prime = 1}^t (1 - \sqrt{2\epsilon_{t^\prime}}) \geq 1 - \sum_{t^\prime}^t \sqrt{2\epsilon_{t^\prime}},
\]
and therefore we have
\[
\|p(\st) - q(\st)\|_1 \leq 2\sum_{t^\prime = 1}^t \sqrt{2\epsilon_{t^\prime}}
\]
This completes the proof.

\subsection{Total Policy Cost Bound}
\label{app:cost_bound}

In this appendix, we use the result above to prove Lemma~\ref{lemma:cost_bound}. This result is based on Ross et al.~\cite{rgb-rilsp-11}, but extends the proof to the case of time-varying finite-horizon systems. We first restate the lemma under the same notation as the previous appendix:
\begin{lemma}
	If $\tv(\trajdist(\st)\| \policy_\params(\st)) \leq 2 \sum_{t=1}^T \sqrt{2\epsilon_t}$, then we can bound the total cost of $\policy_\params$ as
	\[
	\sum_{t=1}^T E_{\policy_\params(\st,\at)}[\cost(\st,\at)] \leq \sum_{t=1}^T \left[ E_{\trajdist(\st,\at)}[\cost(\st,\at)] + 2 \sqrt{\epsilon_t} Q_{\text{max},t}\right],
	\]
	where $Q_{\text{max},t} = \sum_{t^\prime=t}^T \max_{\state_{t^\prime},\action_{t^\prime}}\cost(\state_{t^\prime},\action_{t^\prime})$, the maximum total cost from time $t$ to $T$.
\end{lemma}
We bound the cost of $q$ at time step $t$ according to
\begin{align*}
E_{q(\st, \at)}[\cost(\st, \at)] &= \langle q(\st, \at), \cost(\st, \at) \rangle \\
&= \langle q(\st, \at) - p(\st) q(\at|\st), \cost(\st, \at) \rangle + \langle p(\st) q(\at|\st), \cost(\st, \at) \rangle \\
&= \langle q(\at|\st) [q(\st) - p(\st)], \cost(\st, \at) \rangle + \langle p(\st) [q(\at|\st) - p(\at|\st)], \cost(\st, \at) \rangle + E_{p(\st,\at)}[\cost(\st, \at)] \\
&\leq E_{p(\st,\at)}[\cost(\st, \at)] + \|q(\st) - p(\st)\|_1 \max_{\st, \at} \cost(\st, \at) + \|q(\at|\st) - p(\at|\st)\|_1 \max_{\st, \at} \cost(\st, \at) \\
&\leq E_{p(\st,\at)}[\cost(\st, \at)] + \max_{\st, \at} \cost(\st, \at) \sqrt{2\epsilon_t} + 2\max_{\st, \at} \cost(\st, \at)\sum_{t^\prime = 1}^t \sqrt{2\epsilon_{t^\prime}} 
\end{align*}
If we add up the above quantity over all time $t$, we get
\[
    \sum_{t=1}^T E_{q(\st, \at)}[\cost(\st, \at)] \leq \sum_{t=1}^T E_{p(\st, \at)}[\cost(\st, \at)]
    + \sum_{t=1}^T \sqrt{2\epsilon_t}\max_{\st,\at}\cost(\st,\at)
    + 2\sum_{t=1}^T \max_{\st,\at} \cost(\st,\at) \sum_{t^\prime=1}^t \sqrt{2\epsilon_t^\prime}
\]
which we can rewrite as
\[
\sum_{t=1}^T E_{q(\st,\at)}[\cost(\st,\at)] \leq \sum_{t=1}^T \left[ E_{p(\st,\at)}[\cost(\st,\at)] + \sqrt{2\epsilon_t} \max_{\st, \at} \cost(\st, \at) + 2\sqrt{2\epsilon_t} Q_{\text{max},t} \right]
\]
where $Q_{\text{max},t} = \sum_{t^\prime=t}^T \max_{\state_{t^\prime},\action_{t^\prime}} \cost(\state_{t^\prime},\action_{t^\prime})$.

\end{document}